\documentclass[conference]{IEEEtran}
\IEEEoverridecommandlockouts
\usepackage{cite}
\usepackage{amsmath,amssymb,amsfonts}
\usepackage{algorithmic}
\usepackage{graphicx}
\usepackage{textcomp}
\usepackage{xcolor}

\usepackage{cuted}
\usepackage{amsmath}
\usepackage{hyperref}
\usepackage{pifont}
\usepackage{marvosym}
\usepackage{amssymb}
\usepackage{subcaption}
\usepackage{booktabs}
\usepackage{multirow}
\usepackage{graphicx}
\usepackage{xcolor}
\usepackage{placeins}
\usepackage[table]{xcolor}

\def\BibTeX{{\rm B\kern-.05em{\sc i\kern-.025em b}\kern-.08em
    T\kern-.1667em\lower.7ex\hbox{E}\kern-.125emX}}
\begin{document}

\title{MDE-AgriVLN: Agricultural Vision-and-Language Navigation with Monocular Depth Estimation}


\author{\large Xiaobei Zhao\textsuperscript{\rm 1}, Xingqi Lyu\textsuperscript{\rm 1}, Xin Chen\textsuperscript{\rm 1}\textsuperscript{\Letter}, Xiang Li\textsuperscript{\rm 1}\textsuperscript{\Letter}\thanks{\textsuperscript{\Letter} Corresponding Authors}\thanks{This study is supported by the National Natural Science Foundation of China's Research on Distributed Real-Time Complex Event Processing for Intelligent Greenhouse Internet of Things (Grant No. 61601471).}\normalsize\\ \\
\textsuperscript{\rm 1}China Agricultural University\\
\texttt{\{xiaobeizhao2002,lxq99725\}@163.com, \{chxin,cqlixiang\}@cau.edu.cn}}

\maketitle

\begin{strip}
\centering
\vspace{-1.8cm}
\includegraphics[width=\textwidth]{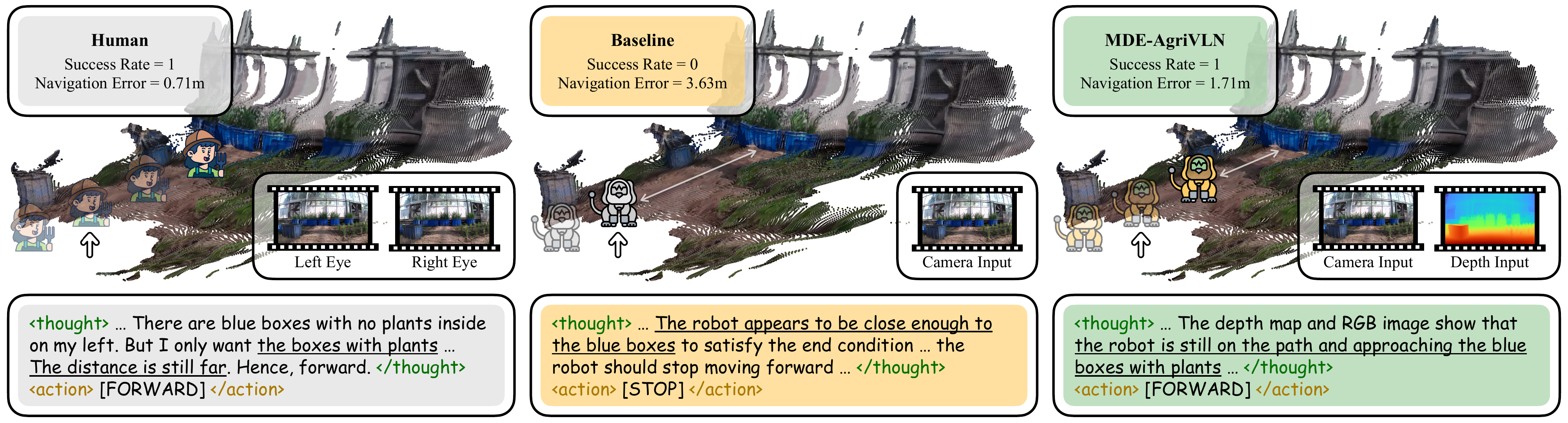}
\captionof{figure}{\small MDE-AgriVLN v.s. Human and Baseline on a representative episode. In every method's section, the right images are the visual inputs at the time step $t = 6.2s$ (marked by \raisebox{-0.15em}{\includegraphics[height=1em]{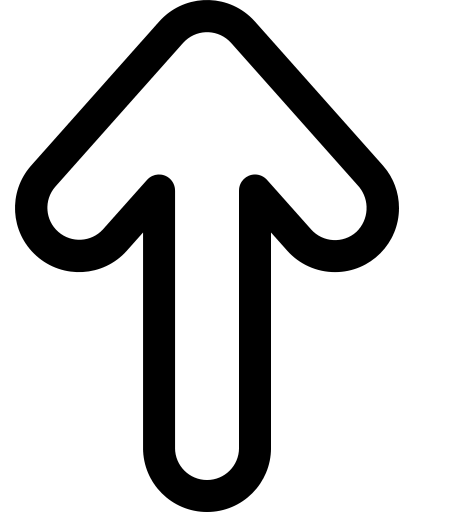}}), the bottom textbox is the reasoning result at the same time step, and the top textbox is the evaluation result. \underline{Underline} marks the pivotal reasoning thoughts.
Instruction: \textit{Now you are standing on the ground of a greenhouse. There are lots of blue boxes in your view, and I need you navigate to the blue box with plants planted on it, which is at the end of the road in front of you. Please go along the path and stop when you are very close to the destination.} (Zoom in to observe more details)}
\label{fig:teaser}
\end{strip}

\begin{abstract}
Agricultural robots are serving as powerful assistants across a wide range of agricultural tasks, nevertheless, still heavily relying on manual operations or railway systems for movement. The AgriVLN method and the A2A benchmark pioneeringly extended Vision-and-Language Navigation (VLN) to the agricultural domain, enabling a robot to navigate to a target position following a natural language instruction. Unlike human binocular vision, most agricultural robots are only given a single camera for monocular vision, which results in limited spatial perception. To bridge this gap, we present the method of Agricultural Vision-and-Language Navigation with Monocular Depth Estimation (MDE-AgriVLN), in which we propose the MDE module generating depth features from RGB images, to assist the decision-maker on multimodal reasoning. When evaluated on the A2A benchmark, our MDE-AgriVLN method successfully increases Success Rate from 0.23 to 0.32 and decreases Navigation Error from 4.43m to 4.08m, demonstrating the state-of-the-art performance in the agricultural VLN domain. 
Code: \href{https://github.com/AlexTraveling/MDE-AgriVLN}{https://github.com/AlexTraveling/MDE-AgriVLN}.
\end{abstract}


\section{Introduction}
\label{sec:introduction}
\par Agricultural robots are serving as powerful assistants across a wide range of tasks, such as laser weeding \cite{COMPAG:LaserWeeding}, growth monitoring \cite{RAL:GrowthMonitoring} and cross-pollination \cite{Cell:Pollination}. 
However, most of them still heavily rely on manual operations or railway systems for movement, resulting in limited mobility and poor adaptability. 
\par In contrast, Vision-and-Language Navigation (VLN) enables agents to follow the natural language instructions to navigate to the target positions \cite{CVPR:R2R,ECCV:VLN-CE}, having demonstrated strong performance across various domains \cite{TMLR}, such as R2R \cite{CVPR:R2R} for indoor room, TOUCHDOWN \cite{CVPR:TOUCHDOWN} for urban street, and AerialVLN \cite{ICCV:AerialVLN} for aerial space. Motivated by prior Vision-Language Model-based approaches \cite{AAAI:NavGPT,ACL:MapGPT,ICRA:Discuss}, Zhao et al. \cite{arXiv:AgriVLN} proposed the AgriVLN method and the A2A benchmark to pioneeringly extend VLN to the agricultural domain, enabling agricultural robots to navigate to the target positions following the natural language instructions. However, when evaluated on the A2A benchmark, AgriVLN only achieves Success Rate (SR) of 0.23 and Navigation Error (NE) of 4.43m, leaving a large gap to the performance of Human. 
\par We attribute this gap to Human's binocular cues \cite{SeeinginDepth} as one of the reasons. There is a distance around 5.5 $\sim$ 7.5 cm between the left and right eyes, leading to a tiny distinction on observation perspective, i.e., the binocular disparity, which can be processed by brain to set up stereoscopic vision. For most agricultural robots, however, the only visual input is a single RGB image, which makes their decision-makers lack the spatial senses, such as on distance. Here we share a simple example to better demonstrate their difference, as illustrated in Fig. \ref{fig:teaser}. At the time step $t = 6.2s$, Baseline only takes a single RGB image as the visual input, then mistakenly thinks that \textit{the robot appears to be close enough to the blue boxes}, resulting in an early $\texttt{STOP}$ with NE = 3.63m. Meanwhile, Human utilizes its stereoscopic vision to perceive \textit{boxes with plants}' depths in a more accurate way, thereby properly thinking that \textit{the distance is still far} then predicting the $\texttt{FORWARD}$ action.
\par To bridge this gap, we present the method of \textbf{Agri}cultural \textbf{V}ision-and-\textbf{L}anguage \textbf{N}avigation with \textbf{M}onocular \textbf{D}epth \textbf{E}stimation (\textbf{MDE-AgriVLN}). First, we propose the MDE module, in which we employ a monocular depth estimator as the backbone and establish two representation paradigms. Second, we integrate the MDE module into the base model to establish our MDE-AgriVLN method. Third, we evaluate it on the A2A \cite{arXiv:AgriVLN} benchmark, successfully increasing SR from 0.23 to 0.32 and decreasing NE from 4.43m to 4.08m. On the example in Fig. \ref{fig:teaser}, MDE-AgriVLN utilizes monocular depth estimation to assist its reasoning, thereby properly thinking that \textit{the robot is still on the path and approaching the blue boxes with plants}, then predicting the $\texttt{FORWARD}$ action, which sufficiently surpasses the performance of Baseline and narrow the performance gap with Human.


\par In summary, our main contributions are as follows:
\begin{itemize}
\item We propose the MDE module based on a monocular depth estimator with two representation paradigms, to generate depth features from monocular RGB images.
\item We propose the MDE-AgriVLN method integrating the MDE module to pioneeringly employ monocular depth estimation to enhance the spatial perception.
\item We implement the ablation experiment proving MDE's effectiveness and the comparison experiment proving MDE-AgriVLN's state-of-the-art performance.
\end{itemize}

\section{Related Works}
\label{sec:related_works}
\subsection{Agricultural Vision-and-Language Navigation}
\par Agriculture-to-Agriculture (A2A) \cite{arXiv:AgriVLN} is a VLN benchmark specially designed for agricultural robots, consisting of 1,560 episodes across 6 types of scene: farm, greenhouse, forest, mountain, garden and village, in which all the instructions belong to the step-by-step format and the action space belongs to the continuous environment.


\subsection{Depth Employment in Vision-and-Language Navigation}
The standard visual input for a VLN method is a panoramic image in the discrete environment \cite{CVPR:R2R} or a front-facing image in the continuous environment \cite{ECCV:VLN-CE} at any time step, which provides abundant semantic informations but limited spatial features. 
To address this issue, some studies \cite{ICME:DG-AdaIN,ICRA:DA-VLN,AAAI:NavGPT} have explored employing the depth feature as an additional modality. 
These methods show the effectiveness of the depth feature, however, they all depend on the depth sensor, which is usually an unaffordable hardware for realistic deployments \cite{ICCV:DepthCamera}. Hence, we are motivated to explore introducing monocular depth estimation to replace the depth sensor. None of the existing methods, as far as we know, have studied on this topic.

\subsection{Monocular Depth Estimation}
The purpose of monocular depth estimation is to transform a single photographic image into a depth map \cite{NeurIPS:DepthMapPrediction}. Previous studies have explored the improvement from various fronts, such as piecewise planarity prior \cite{CVPR:P3Depth}, neural conditional random fields \cite{CVPR:FC-CRFs} and first-order variational constraints \cite{ICLR:VA-DepthNet}. We attempt employing three state-of-the-art models \cite{NeurIPS:DepthAnythingV2,NeurIPS:Pixel-Perfect-Depth,ICLR:Depth-Pro} as the backbone for our MDE module, then implement the ablation experiment to further select the most compatible one.
\par The complete version of Related Works with more details is available in Appendix.

\begin{figure*}[t]
\centering
\includegraphics[width=1.0\linewidth]{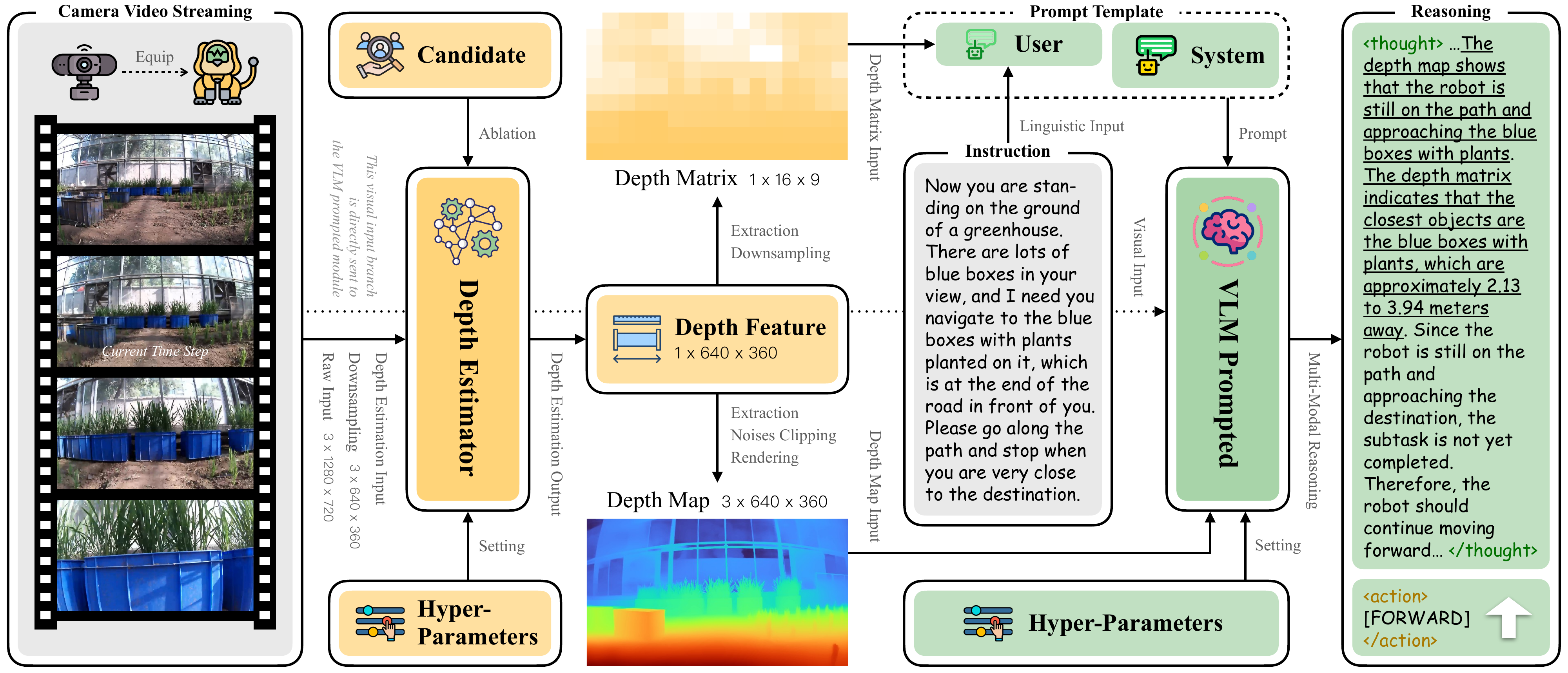}
\caption{\small MDE-AgriVLN methodology illustration: The MDE module (yellow part) takes a single frame from a camera video streaming as input, to output the depth feature in two classifications of representation. The base model (green part) simultaneously understand the instruction, RGB input and depth input, to reason the most appropriate low-level action with an explicit thought.}
\label{fig:method}
\end{figure*}

\section{Methodology}
\label{sec:methodology}
In this section, we present the MDE-AgriVLN method, as illustrated in Figure \ref{fig:method}. First, we introduce the task definition in Sec. \ref{sec:task_definition}. Second, we present the module of \textbf{M}onocular \textbf{D}epth \textbf{E}stimation (\textbf{MDE}) in Sec. \ref{sec:MDE}. Third, we integrate it into the base model to establish MDE-AgriVLN in Sec. \ref{sec:base_model}.

\subsection{Task Definition}
\label{sec:task_definition}
The task of Agricultural Vision-and-Language Navigation \cite{arXiv:AgriVLN} is defined as follows: In each episode, the model is given an instruction in natural language, denoted as $W = \langle w_1, w_2, \dots, w_L \rangle$, where $L$ is the number of words. At each time step $t$, the model is given the front-facing RGB image $I_t$. The purpose is understanding both $W$ and $I_t$, to select the best low-level action $\hat{a_t}$ from action space $\{ \texttt{FORWARD}$, $\texttt{LEFT ROTATE}$, $\texttt{RIGHT ROTATE}$, $\texttt{STOP} \}$, thereby leading the robot navigate from the starting point to the destination. 

\subsection{Monocular Depth Estimation Module}
\label{sec:MDE}
At each time step $t$, the visual input is an RGB image $I_t \in \mathbb{R}^{3 \times 720 \times 1280}$. 
In the pre-processing stage, we downsample it to $I'_t \in \mathbb{R}^{3 \times \mathbb{H} \times \mathbb{W}}$, where $\mathbb{H}$ and $\mathbb{W}$ are set to 360 and 640 as the default values, respectively. We employ the monocular depth estimator $\mathcal{E}\,(\,\cdot\,)$ to estimate the depth $D_t$, defined as:
\begin{equation}
D_t = \mathcal{E}_{\phi}(I'_t), \quad D_t \in \mathbb{R}^{1 \times \mathbb{H} \times \mathbb{W}}
\end{equation}
where $D_t$ is saved in the numpy npz format. $\phi$ is the hyper-parameter of $\mathcal{E}\,(\,\cdot\,)$. 
Then we establish two representation paradigms for $D_t$:
\subsubsection{Depth Matrix}
A complete depth matrix is massive and tedious. Therefore, if we directly inputted a complete depth matrix to the Vision-Language Model (VLM) in the base model, the context length would exceed the limitation. To address this issue, we innovate a simple yet effective downsampling function, denoted as $\mathcal{S}\,(\,\cdot\,)$. We denote the downsampling ratio as $r$, then the downsampled depth matrix $D_t^{matrix}$ is in the size of $\mathbb{H} / r \times \mathbb{W} / r$, denoted as $\mathbb{H}' \times \mathbb{W}'$. We uniformly split a complete frame into $\mathbb{H}' \times \mathbb{W}'$ dispersed patches, then only retain the central single pixel in each patch. The complete downsampling process is defined as:
\begin{equation}
D_t^{matrix} = \mathcal{S}(D_t, r)
\end{equation}
where every element in $D_t^{matrix}$ is in absolute metric scale, i.e., in meters. $r$ is set to 40 as the default value.
\subsubsection{Depth Map}
To avoid the extreme noises, we employ percentile-based clipping to replace all the depth values smaller than the $l$ percentile and bigger than the $u$ percentile with the threshold values, defined as:
\begin{equation}
D_t' = \mathcal{C}\big(D_t, \, \mathcal{P}_l(D_t), \, \mathcal{P}_{u}(D_t)\big)
\end{equation}
where $\mathcal{C}\,(\,\cdot\,)$ is the clipping function. $\mathcal{P}_l(\,\cdot\,)$ and $\mathcal{P}_{u}(\,\cdot\,)$ are the lower and upper clipping thresholds, respectively. $l$ and $u$ are set to 1 and 99 as the default values, respectively. We normalize $D_t'$ to the range [0, 1], defined as:
\begin{equation}
D_t^n = \frac{D_t' - \min(D_t')}{\max(D_t') - \min(D_t')}
\end{equation}
Then we employ the colormap function from Matplotlib \cite{Matplotlib} to render it to the depth map $D_t^{map}$ in the pseudo-color paradigm, defined as:
\begin{equation}
D_t^{map} = \mathcal{R}_m(D_t^n)
\end{equation}
where $\mathcal{R}_m\,(\,\cdot\,)$ is the rendering function. $m$ is the color mode, and we set it to \textit{turbo reversed}, in which objects are rendered from red to blue according to their distance from near to far.

\subsection{Base Model}
\label{sec:base_model}
We follow the VLM-based architecture of AgriVLN \cite{arXiv:AgriVLN} as our base model, denoted as $VLM\,(\,\cdot\,)$, in which the MDE module is integrated between the visual input layer and the multimodal reasoning layer. At each time step $t$, $VLM\,(\,\cdot\,)$ simultaneously  understand the linguistic features - $W$ - and the visual features - $I'_t$, $D_t^{matrix}$ and $D_t^{map}$ - to reason the most appropriate low-level action $\hat{a_t}$, defined as:
\begin{equation}
\begin{aligned}
\hat{a_t}, \rho_t &= VLM(p_s, p_{u, t}, I'_t, D_t^{map}) \\
p_{u, t} &= W \; \Vert \; D_t^{matrix}
\end{aligned}
\end{equation}
where $p_s$ and $p_{u, t}$ are the system and user prompts, respectively. $\rho_t$ is the reasoning thought, providing an explicit interpretation. $\Vert$ is the literal concatenation. In an episode, $VLM\,(\,\cdot\,)$ ends when one of the following conditions happens: 
\begin{quote}
1) $\hat{a_{t'}}$ = $\texttt{STOP}$; \\
2) The predicted action sequence $\langle \hat{a_{t'-\tau}}, \hat{a_{t'-\tau+1}},$ $\dots, \hat{a_{t'}} \rangle$ is deviated to the ground-truth action sequence $\langle a_{t'-\tau}, a_{t'-\tau+1}, \dots, a_{t'} \rangle$; \\
3) $t'$ reaches the max limitation of time step.
\end{quote}
where $t'$ is the stopping time step. $\tau$ is the deviation time threshold, which is set to 4s following A2A \cite{arXiv:AgriVLN}'s setting.

\begin{table*}[t]
\caption{\small Comparison experiment results between MDE-AgriVLN and state-of-the-art methods on the low-complexity portion (subtask $=$ 2), the high-complexity portion (subtask $\geq$ 3), and the whole of the A2A benchmark.}
\label{tab:comparison_experiment}
\begin{center}
\resizebox{\linewidth}{!}{
\renewcommand{\arraystretch}{1.1}
\begin{tabular}{rlccccc|cccc|cccc}
\toprule
\multirow{2}{*}{\textbf{\#}} & \multirow{2}{*}{\textbf{Method}} & \multirow{2}{*}{\textbf{MDE Representation}} & 
\multicolumn{4}{c}{\textbf{A2A ( low-complexity )}} & 
\multicolumn{4}{c}{\textbf{A2A ( high-complexity )}} & 
\multicolumn{4}{c}{\textbf{A2A}} \\
\cmidrule(lr){4-7} \cmidrule(lr){8-11} \cmidrule(lr){12-15}
& & & \textbf{SR}$\uparrow$ & \textbf{NE}$\downarrow$ & \textbf{TU}$_\text{p}$$\downarrow$ & \textbf{TU}$_\text{c}$$\downarrow$ &
\textbf{SR}$\uparrow$ & \textbf{NE}$\downarrow$ & \textbf{TU}$_\text{p}$$\downarrow$ & \textbf{TU}$_\text{c}$$\downarrow$ & 
\textbf{SR}$\uparrow$ & \textbf{NE}$\downarrow$ & \textbf{TU}$_\text{p}$$\downarrow$ & \textbf{TU}$_\text{c}$$\downarrow$ \\
\midrule
1 & Random   & -                       & 0.13 & 7.30 & - & - & 0.04 & 6.74 & - & - & 0.09 & 7.03 & - & - \\
2 & Fixed    & -                       & 0.00 & 0.00 & - & - & 0.06 & 6.32 & - & - & 0.03 & 3.06 & - & - \\
\midrule \rowcolor{gray!19} \multicolumn{15}{c}{\textit{State-of-the-Art}} \\
3 & SIA-VLN     & -            & \underline{0.26} & 2.92             & \textbf{1,244}    & \textbf{158}    & \underline{0.02} & \underline{6.24} & 1,327             & 177             & \underline{0.15} & \underline{4.52} & \textbf{1,284}    & 167 \\
4 & DILLM-VLN   & -            & 0.31             & \textbf{2.52}    & 1,322             & 189             & 0.10             & 5.56             & \textbf{1,279}    & \textbf{132}    & 0.21             & \textbf{3.99}    & 1,301             & 162 \\
5 & AgriVLN     & -            & 0.30             & \underline{3.72} & 1,261             & 163             & 0.16             & 5.18             & 1,401             & 144             & 0.23             & 4.43             & 1,329             & \textbf{154} \\
\midrule \rowcolor{gray!19} \multicolumn{15}{c}{\textit{Ours}} \\
6 & MDE-AgriVLN & Depth Matrix & \textbf{0.44}    & 3.05             & 2,156             & 197             & 0.19             & 5.17             & 2,257             & \underline{199} & \textbf{0.32}    & 4.08             & 2,205             & \underline{198} \\
7 & MDE-AgriVLN & Depth Map    & 0.40             & 3.70             & 1,622             & \underline{204} & \textbf{0.23}    & \textbf{4.81}    & 1,739             & 170             & \textbf{0.32}    & 4.24             & 1,679             & 187 \\
8 & MDE-AgriVLN & Hybrid       & 0.41             & 3.57             & \underline{2,545} & 193             & 0.21             & 5.10             & \underline{2,585} & 194             & 0.31             & 4.31             & \underline{2,565} & 194 \\
\midrule
9 & Human    & -                       & 0.93 & 0.32 & - & - & 0.80 & 0.82 & - & - & 0.87 & 0.57 & - & - \\
\bottomrule
\end{tabular}
}

\footnotesize 
\vspace{2.0ex}
\textbf{Bold} and \underline{underline} mark the best and worst scores, respectively.

\end{center}
\end{table*}

\section{Experiments}

\subsection{Experimental Settings}
We implement all the experiments on the A2A \cite{arXiv:AgriVLN} benchmark. In the MDE-AgriVLN method, we locally deploy DeepSeek-R1-32B\footnote{DeepSeek-R1-32B is the distilled version based on the Qwen2 architecture.} \cite{Nature:DeepSeek} with Q4\_K\_M quantization as the Large Language Model, Qwen2.5-VL-32B \cite{arXiv:Qwen} with Q4\_K\_M quantization as the VLM, and Depth Pro \cite{ICLR:Depth-Pro} as the monocular depth estimator. In Qwen2.5-VL-32B, we set the inference temperature to 1e-4 to ensure the response stability. All the experiments run on a single NVIDIA L20 GPU with 48G video memory.

\subsection{Evaluation Metrics}
We follow two standard VLN evaluation metrics \cite{CVPR:R2R}: Success Rate (SR) and Navigation Error (NE). NE measures the path length between the stopping position and the target position. SR measures the rate successfully reaching the target position within a 2-meter NE. Besides, we further introduce Token Usage (TU$_\text{p}$ and TU$_\text{c}$) to evaluate VLM's average reasoning burden at a single time step, calculated by:
\begin{equation}
TU_p = \frac{1}{|D_p|}\sum_{\tau_p\in D_p} \tau_p, \qquad
TU_c = \frac{1}{|D_c|}\sum_{\tau_c\in D_c} \tau_c
\end{equation}
where $\tau_p$ is the prompt token usage at a single time step. $D_p$ is the noise-trimmed set of prompt token usages. $TU_p$ is the average token usage in the prompt stage. For the completion stage, $\tau_c$, $D_c$ and $TU_c$ are defined in an analogous manner.

\subsection{Comparison Experiment}
\par We compare MDE-AgriVLN with three state-of-the-art methods: SIA-VLN \cite{EMNLP:SIA-VLN}, DILLM-VLN \cite{RAL:DILLM-VLN} and AgriVLN \cite{arXiv:AgriVLN}. Besides, the methods of Random, Fixed and Human are reproduced as the lower and upper bounds, respectively. 
The results are shown in Table \ref{tab:comparison_experiment}, in which we select the depth matrix-based representation version (\#6) as our primary method (will be discussed in Sec. \ref{sec:ablation_experiment_depth_representation}). On the whole A2A benchmark, MDE-AgriVLN (\#6) achieves SR of 0.32 and NE of 4.08m, successfully increasing SR by 9 percentage points and decreasing NE by 0.35m compared to the AgriVLN baseline (\#5), proving that the depth features from the MDE module can be effectively utilized by the VLM-based decision-maker to improve its spatial perception. 
Besides, MDE-AgriVLN surpasses all the baselines on both low-complexity and high-complexity portions of A2A, demonstrating the state-of-the-art performance in the agricultural VLN domain.

\subsection{Ablation Experiment}
\label{sec:ablation_experiment}

\begin{table}[t]
\caption{\small Ablation experiment results on monocular depth estimator.}
\label{tab:ablation_experiment_monocular_depth_estimation_model}
\begin{center}
\resizebox{\linewidth}{!}{
\renewcommand{\arraystretch}{1.1}
\begin{tabular}{rcccccc}
\toprule
\# & \textbf{Model} & \textbf{Depth} & \textbf{SR}$\uparrow$ & \textbf{NE}$\downarrow$ & \textbf{TU}$_\text{p}$$\downarrow$ & \textbf{TU}$_\text{c}$$\downarrow$ \\ 
\midrule
10 & Depth Anything V2   & M. & \underline{0.30} & \underline{4.23} & \underline{2,287} & \underline{222} \\
11 & Pixel-Perfect Depth & R. & \textbf{0.34}    & 4.19             & \textbf{1,626}    & 212 \\
12 & Depth Pro           & M. & 0.32             & \textbf{4.08}    & 2,205             & \textbf{198} \\
\bottomrule
\end{tabular}
}

\footnotesize 
\vspace{2.0ex}
\textbf{Bold} and \underline{underline} mark the best and worst scores, respectively. \\ M. and R. represent metric and relative depths, respectively.

\vspace{-0.1cm}

\end{center}
\end{table}

\begin{figure}[t]
\centering
  \begin{subfigure}[b]{0.49\linewidth}
    \includegraphics[width=\linewidth]{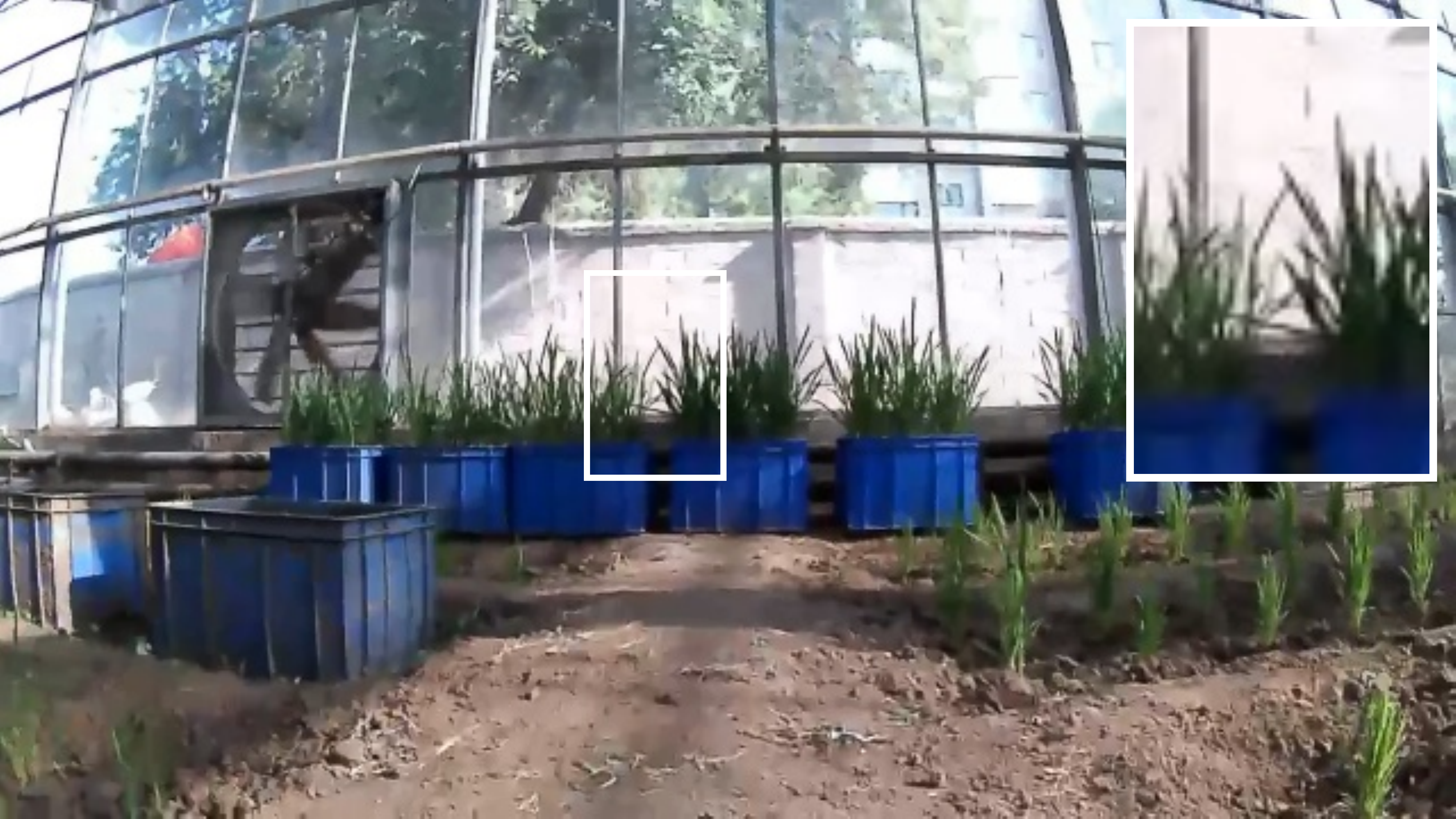}
    \caption{Input}
    \vspace{1.0ex}
  \end{subfigure}
  \hfill
  \begin{subfigure}[b]{0.49\linewidth}
    \includegraphics[width=\linewidth]{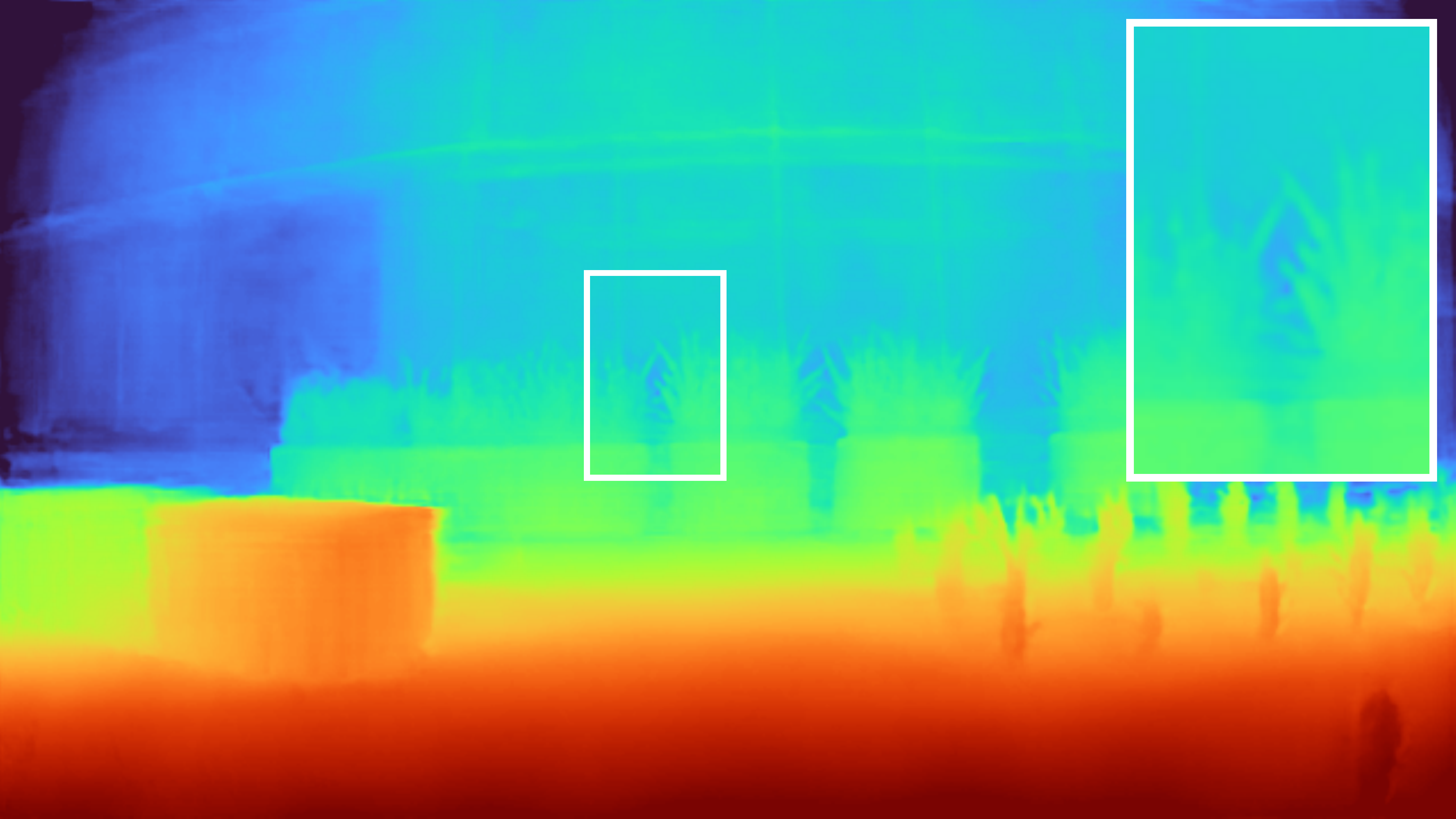}
    \caption{Depth Anything V2}
    \vspace{1.0ex}
  \end{subfigure}
  \begin{subfigure}[b]{0.49\linewidth}
    \includegraphics[width=\linewidth]{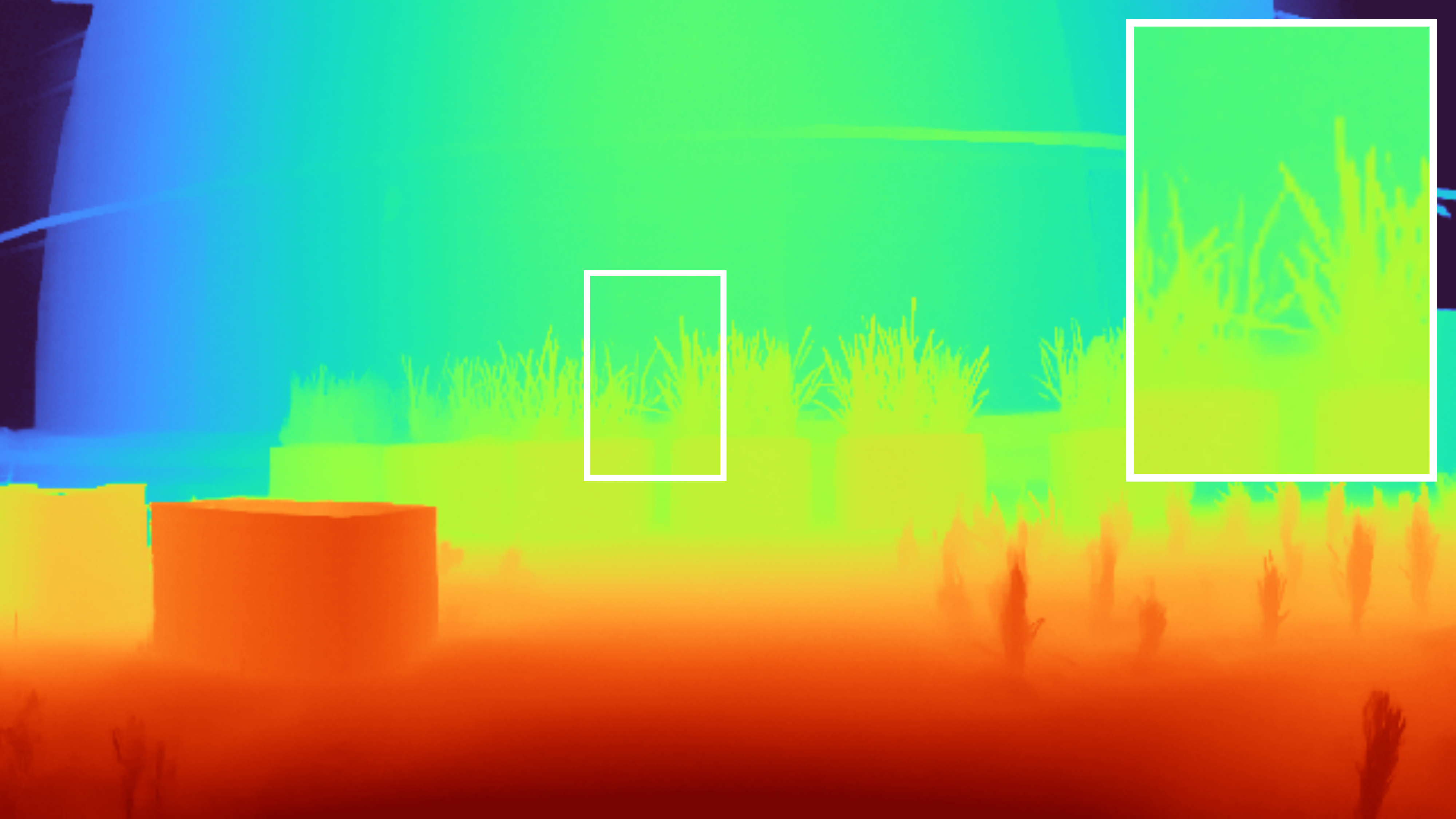}
    \caption{Pixel-Perfect Depth}
  \end{subfigure}
  \hfill
  \begin{subfigure}[b]{0.49\linewidth}
    \includegraphics[width=\linewidth]{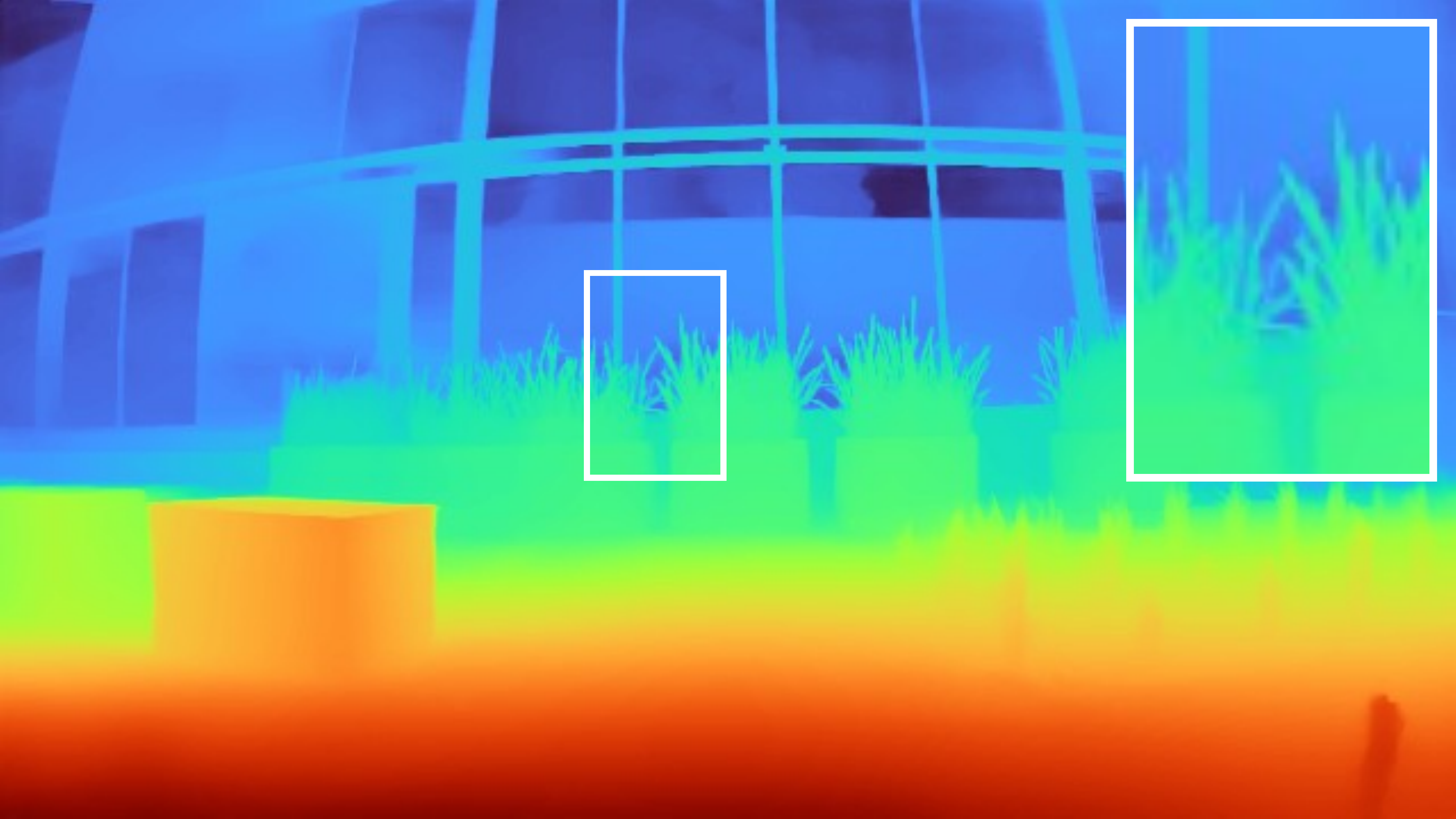}
    \caption{Depth Pro}
  \end{subfigure}
\caption{\small Ablation experiment illustration on monocular depth estimator. (Zoom in to observe more details)}
\vspace{-0.15cm}
\label{fig:ablation_experiment_illustration}
\end{figure}

\begin{figure*}[t]
\centering
\includegraphics[width=1.0\linewidth]{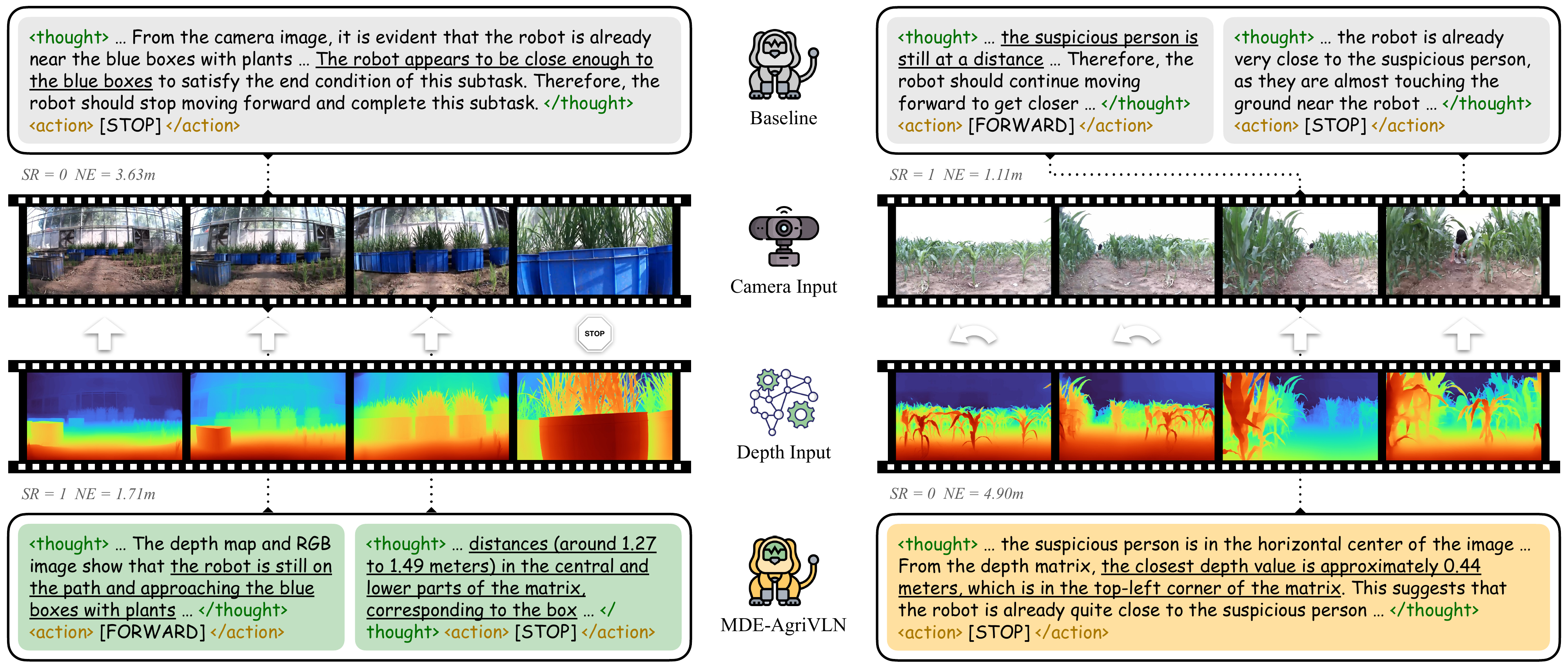}
\caption{\small Qualitative experiment illustration: Baseline reasoning results, camera inputs, ground-truth actions, depth inputs and MDE-AgriVLN reasoning results are shown from top to bottom, respectively. \underline{Underline} marks the pivotal reasoning thoughts. The successful example's instruction (left): \textit{Now you are standing on the ground of a greenhouse. There are lots of blue boxes in your view, and I need you navigate to the blue box with plants planted on it, which is at the end of the road in front of you. Please go along the path and stop when you are very close to the destination.} The failed example's instruction (right): \textit{There is a suspicious person in the farm. We need to inquire about her situation. First, you need left ratate until she is in your horizontal center. Then, walk towards her. Stop when you are very very close to her.}}
\label{fig:qualitative_experiment}
\end{figure*}

\subsubsection{Depth Representation}
\label{sec:ablation_experiment_depth_representation}
We ablate the depth representation paradigm in the MDE module, as shown in Table \ref{tab:comparison_experiment}. We are surprised that either the depth matrix-only version (\#6) or the depth map-only version (\#7) performs even a little better than the hybrid version (\#8), because the hybrid version provides a more comprehensive depth representation. We attribute this interesting phenomenon to the semantic integrality and alignment burden. Specifically, we carefully inspect and compare the reasoning thoughts across different representation paradigms, then we find that either a matrix or a map can already sufficiently represent an estimated depth, therefore, the hybrid of them cannot provide much more semantic. In contrast, the alignment between a RGB image, a depth matrix and a depth map occupies large reasoning attention, which inevitably distracts the decision making. Considering both accuracy and consumption, we select the depth matrix-only version as our primary method.

\subsubsection{Monocular Depth Estimator}
\par We ablate Depth Pro \cite{ICLR:Depth-Pro} with the other two models, Depth Anything V2 \cite{NeurIPS:DepthAnythingV2} and Pixel-Perfect Depth \cite{NeurIPS:Pixel-Perfect-Depth}, as shown in Table \ref{tab:ablation_experiment_monocular_depth_estimation_model} and Fig. \ref{fig:ablation_experiment_illustration}. We find that Depth Anything V2's depth estimation accuracy is relatively worse, e.g., it confuse the target plant box and the greenhouse background wall to almost same metric depths in the example, which tends to provide negative effects to the decision-maker. Pixel-Perfect Depth can only estimate in a relative scale, but brings an excellent VLN performance with the slightest TU$_\text{p}$ of 1,626. We attribute its success to the consistency across different scale objects, as illustrated in the example, in which other models accurately estimate the left big box but struggle with the right small seedlings, while Pixel-Perfect Depth can accurately estimate both of them. However, the lack of metric value makes its NE tends to be farther. In contrast, Depth Pro utilizes its relatively accurate metric depths to help moving closer to a target position with even less reasoning burden, achieving the best NE of 4.08m and TU$_\text{c}$ of 198. Considering both accuracy and stability, we select Depth Pro as the monocular depth estimator in our MDE module.

\begin{table}[!t]
\caption{\small Ablation experiment results on scene classification.}
\label{tab:ablation_experiment_scene_classification}
\begin{center}
\resizebox{\linewidth}{!}{
\renewcommand{\arraystretch}{1.1}
\begin{tabular}{rccccc}
\toprule
\# & \textbf{Scene Classification} & \textbf{Depth Input} & \textbf{SR}$\uparrow$ & \textbf{NE}$\downarrow$ \\ 
\midrule
13 & \multirow{2}{*}{Farm}       & Depth Matrix from MDE & \textbf{0.60} & \textbf{2.32} \\
14 &                             & $\times$              & 0.47          & 2.40          \\
\midrule
15 & \multirow{2}{*}{Greenhouse} & Depth Matrix from MDE & \textbf{0.19} & 3.45          \\
16 &                             & $\times$              & 0.12          & \textbf{3.37} \\
\midrule
17 & \multirow{2}{*}{Forest}     & Depth Matrix from MDE & \textbf{0.13} & \textbf{5.04} \\
18 &                             & $\times$              & 0.06          & 5.65          \\
\midrule
19 & \multirow{2}{*}{Mountain}   & Depth Matrix from MDE & \textbf{0.36} & \textbf{4.25} \\
20 &                             & $\times$              & 0.24          & 4.68          \\
\midrule
21 & \multirow{2}{*}{Garden}     & Depth Matrix from MDE & \textbf{0.30} & \textbf{5.67} \\
22 &                             & $\times$              & 0.23          & 6.48          \\
\midrule
23 & \multirow{2}{*}{Village}    & Depth Matrix from MDE & \textbf{0.33} & \textbf{4.11} \\
24 &                             & $\times$              & 0.27          & 4.12          \\
\bottomrule
\end{tabular}
}

\footnotesize 
\vspace{2.0ex}
\textbf{Bold} marks the better score in every scene classification. \\ The $\times$ symbol represents “without”.

\vspace{-0.15cm}

\end{center}
\end{table}

\subsubsection{Scene Classification}
\par We further remove the proposed MDE module to analyze its effect on different scene classifications, as shown in Table \ref{tab:ablation_experiment_scene_classification}. When MDE is removed, the performance sharply drops across all the six scene classifications, especially in the challenging forest scene (\#17, \#18), where SR drops over by more than half, demonstrating its essential effectiveness. Nevertheless, in some scene classifications, we note relatively slight performance losses, such as in farm (\#13, \#14), where SR drops about just 20\% and NE increases just 0.08m. We attribute this large distinction to the different chaotic degree. In a farm, most of the crops are planted with similar sizes in an organized way. In a forest, however, all the plants grow with diversified sizes in a basically natural way. Consequently, an image from forest tends to be much more chaotic than an image from farm. Thus, the decision-maker relies more on the depth features in a forest than in a farm. 

\subsection{Qualitative Experiment}
\label{sec:qualitative_experiment}
\par Towards a more comprehensive discussion, we share both a successful example and a failed example, as illustrated in Fig. \ref{fig:qualitative_experiment}. We mark all the VLM's pivotal reasoning thoughts with \textit{italics} in the text and \underline{underline} in the figure. 
The demonstration video is available in Appendix.
\par In the successful example, at the time step $t = 6.2s$, Baseline thinks that \textit{the robot appears to be close enough to the blue boxes}, so decides to $\texttt{STOP}$, leaving a 3.63m navigation error. At the same time step, the depth features assists MDE-AgriVLN to more accurately recognize the distance that \textit{the robot is still on the path and approaching the blue boxes with plants}, effectively leading to keep going $\texttt{FORWARD}$. At the time step $t = 10.4s$, MDE-AgriVLN aligns \textit{the box} with \textit{the central and lower parts of the matrix}, which indicates an \textit{around 1.27-to-1.49-meters distance}, so successfully predicts the $\texttt{STOP}$ action with only an 1.71m navigation error. From this successful example, we suggest that the depth features can effectively improve VLM's spatial perception, and the MDE module can provide them in a reliable accuracy.
\par In the failed example, both methods smoothly pass the $\texttt{LEFT ROTATE}$ step. At the time step $t = 6.4s$, Baseline thinks that \textit{the suspicious person is still at a distance}, so properly keeps going $\texttt{FORWARD}$. However, MDE-AgriVLN predicts the wrong $\texttt{STOP}$ action, because it observes that \textit{the closest depth value is approximately 0.44 meters}, then mistakenly aligns it with the person, while this depth actually belongs to \textit{the top-left corner}, i.e., a plant of corn. From this failed example, we suggest that the alignment between camera image and depth image is a big challenge for VLM, and a mistaken alignment may results in a disastrous misleading.

\section{Conclusion}
\label{sec:conclusion}
In this paper, we present the MDE-AgriVLN method, effectively improving SR and NE from 0.23 and 4.43m to 0.32 and 4.08m on the A2A \cite{arXiv:AgriVLN} benchmark, respectively, demonstrating the state-of-the-art performance in the agricultural VLN domain.
\par During the experiments, we also find two main weaknesses: 1) As mentioned in Sec. \ref{sec:qualitative_experiment}, MDE-AgriVLN occasionally misalign an object between in the camera image and in the depth image. 2) The time cost of the MDE module on a single time step is about 0.2s $\sim$ 0.8s. Hence, we suggest a further improvement on reasoning stability and real-time capability.
\par In the future, in addition to the above improvement on the existing weaknesses, we plan to explore how an agricultural VLN method would perform if an instruction itself had errors, and propose a new mechanism to promote the robustness. 

\section*{Acknowledgment}
Thanks to Kota Kinabalu, Brunei, Penang and Siem Reap for the impressive traveling experiences, giving us a chilled vibe for writing. Thanks to Kathy Wang for being the participant offering the Human reasoning thoughts in Fig. \ref{fig:teaser}. Thanks to Yuanquan Xu, the inspiration to us.

\bibliographystyle{IEEEbib}
\bibliography{icme2026references}


\end{document}